\documentclass[runningheads]{llncs}

\usepackage[T1]{fontenc}
\usepackage{graphicx}
\usepackage{booktabs}
\usepackage{amsmath}
\usepackage{multirow}
\usepackage{siunitx}
\begin{document}

\title{SignMAE: Segmentation-Driven Self-Supervised Learning for Sign Language Recognition}

\titlerunning{SignMAE}

\author{Kunyuan Xie\orcidID{0009-0000-7660-8192} \and
Zhixi Cai\orcidID{0000-0001-7978-0860}\thanks{Corresponding author} \and
Kalin Stefanov\orcidID{0000-0002-0861-8660}}

\authorrunning{Xie, K. et al.}

\institute{Monash University, Australia\\
\email{kunyuan.xie@monash.edu}\\
\email{zhixi.cai@monash.edu}\\
\email{kalin.stefanov@monash.edu}
}

% typeset the header of the contribution
\maketitle

\begin{abstract}
Subtle hand differences make sign language recognition challenging, yet many existing methods rely on encoders pretrained on generic action datasets that poorly capture such fine-grained cues. 
We propose a self-supervised pretraining method for sign language recognition that uses segmentation-based masking to adapt to the presence and motion of key body parts, rather than treating hand poses as static visual tokens. 
The resulting mask-and-reconstruct objective improves fine-grained sign representation learning. 
On WLASL, NMFs-CSL, and Slovo, our encoder achieves state-of-the-art performance, improving per-instance and per-class Top-1 accuracy while using fewer input frames and modalities than comparable encoders.

\end{abstract}
\section{Introduction}
\label{sec:introdcution}
Sign languages are visual-gestural languages used by Deaf communities worldwide. Unlike spoken languages, they convey meaning through coordinated movements of the hands, body, and face. Isolated sign language recognition (ISLR) classifies short video clips into single-word glosses and underpins downstream tasks such as sign spotting, retrieval, translation, and continuous sign language recognition.

In ISLR, the hands carry most of the linguistic signal, yet they occupy only a small portion of each frame and appear under varying backgrounds and clothing. Existing methods mainly use raw video or keypoints extracted from video. Skeleton-based approaches model hand keypoints alone or together with body keypoints, but standard keypoint extractors often fail, and keypoint representations omit fine-grained visual cues. Although self-supervised mask-and-reconstruct methods have been applied to sign keypoints~\cite{zhao2023best,hu2021signbert}, video-based models still outperform them.

Most video-based ISLR methods~\cite{zuo2023natural,li2020wordlevelsignlanguage,Hosain_2021_WACV} use backbones pretrained for generic action recognition, which process the entire frame and may under-emphasize the hands. VideoMAEv2~\cite{wang2023videomaev2scalingvideo} demonstrates the effectiveness of self-supervised video pretraining by masking spatio-temporal tubes and reconstructing missing content. However, its random tube masking is not well suited to sign language, since many sampled tubes contain little useful hand information.

Despite recent progress, ISLR remains challenging. On WLASL~\cite{li2020wordlevelsignlanguage}, which contains 2,000 glosses, top-1 accuracy is still only around 60\%. This gap motivates more effective hand-centric representation learning. To this end, we propose SignMAE, a segmentation-guided self-supervised framework that learns complementary representations from video and keypoints. During pretraining, modality-specific encoder-decoder backbones reconstruct only tokens from hand regions, encouraging attention to linguistically salient cues. We further design masking strategies tailored to the hand regions. For downstream ISLR, the video and keypoint encoders are finetuned separately and fused through cross-attention.

Our contributions are as follows:
\begin{itemize}
\item We propose SignMAE, a self-supervised framework that learns complementary hand-centric representations from video and keypoints for ISLR.
\item We introduce segmentation-guided masking strategies that reconstruct only hand-region tokens during pretraining, encouraging both encoders to focus on linguistically salient information.
\item We validate SignMAE on NMFs-CSL~\cite{Hu_2021}, Slovo~\cite{Kapitanov_2023}, and WLASL~\cite{li2020wordlevelsignlanguage}, achieving state-of-the-art performance with fewer input frames and modalities than prior methods.
\end{itemize}

\section{Related Work}
\label{sec:related_work}
Early isolated sign language recognition methods typically use either keypoints or video alone. Keypoint-based approaches, including ST-GCN variants and more recent self-supervised pose models, capture spatio-temporal structure efficiently but are limited by keypoint detection errors and the loss of fine-grained visual detail~\cite{hu2021signbert,zhao2023best,zhao2024masamotionawaremaskedautoencoder,lee2023humanpartwise3dmotion,Bohacek_2022_WACV}. Video-based methods learn directly from RGB frames using 2D CNN-RNN pipelines, 3D CNNs, or transformer-style backbones, and generally achieve stronger performance, though most rely on architectures developed for generic action recognition rather than sign-specific representation learning~\cite{li2020wordlevelsignlanguage,carreira2018quovadisactionrecognition,Hosain_2021_WACV,Hu_2021}. Multi-modal methods combine video and keypoints, or additional modalities such as optical flow and depth, to improve robustness, but often require longer input clips, heavier architectures, or more modalities during inference~\cite{Vazquez-Enriquez_2021_CVPR,hu2021signbert,Jiang_2021_CVPR,zuo2023natural}. More broadly, masked autoencoding has become an effective self-supervised paradigm for visual representation learning: VideoMAE ~\cite{tong2022videomaemaskedautoencodersdataefficient} extends this idea to videos with random tube masking, and VideoMAEv2~\cite{wang2023videomaev2scalingvideo} improves efficiency with a stronger training recipe and partial masked-token decoding. However, such generic masking strategies are not optimized for sign language, where hands occupy only a small portion of the frame. This motivates our segmentation-guided, hand-centric pretraining method, which focuses reconstruction on linguistically informative hand and arm regions while learning complementary video and keypoint representations for ISLR.

\section{Methodology}
\label{sec:methodology}
Our pipeline proceeds through four sequential stages: 1) Data preprocessing extracts important patches that contain the hand regions for downstream learning; 2) Uni-modal self‑supervised pretraining is performed independently on each modality: two video streams and one keypoint stream; 3) Uni‑modal finetuning likewise treats the two modalities separately: the reconstruction decoder is replaced by a classification head, and each encoder is supervised to predict the gloss label of its input video; and 4) Multi‑modal fusion uses the fine-tuned frozen encoders and integrates their representations with a cross‑attention module, allowing joint perception over appearance and motion.

The effectiveness of SignMAE, like other ISLR approaches, depends on the quality of intermediate outputs produced by off-the-shelf detectors for bounding boxes, segmentation, and keypoints.
Challenging conditions, such as self-occlusion, low lighting, or motion blur, can degrade these signals and consequently affect downstream performance.
This limitation is not unique to our framework: all prior ISLR methods that incorporate keypoint extraction similarly rely on pretrained models for these steps.
Importantly, our pipeline is designed to maximise the signal-to-noise ratio for pretraining and finetuning, not to eliminate every possible imperfection.
Small, isolated inaccuracies from the detectors typically introduce negligible noise relative to the full sequence, and the global motion pattern of a sign remains intact.
As a result, minor upstream errors rarely compromise the quality of the learned representations, while broadly reliable preprocessing remains sufficient for SignMAE to achieve strong performance.
More details on the data preprocessing pipeline are provided in the Supplementary Material.

\begin{figure*}[t]
\centering
\includegraphics[width=1\textwidth]{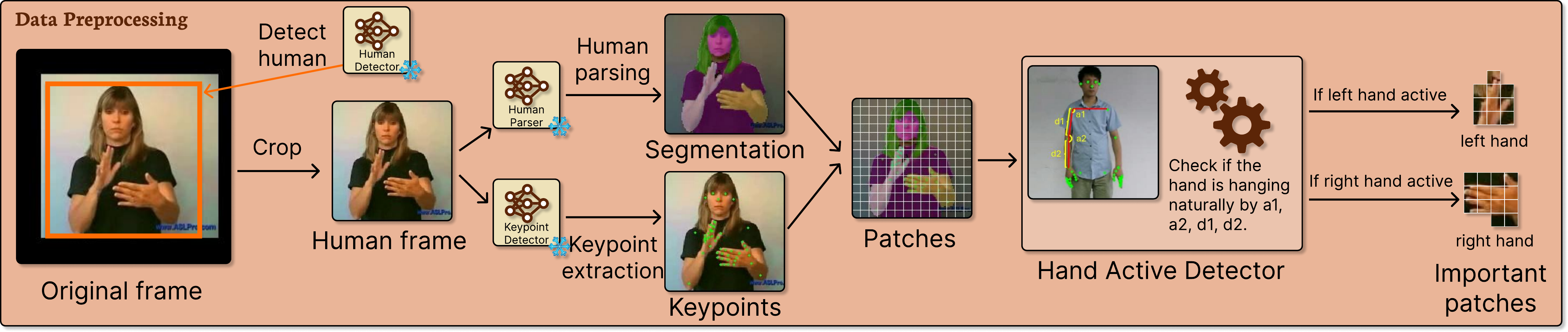}
\caption{Illustration of the data preprocessing pipeline for our SignMAE framework. The data preprocessing pipeline extracts important patches and checks hand movement based on body segments and keypoints extracted from pretrained models.}
\label{fig:data_preprocessing}
\end{figure*}

\subsection{Data Preprocessing}
To maximise the signal-to-noise ratio for later stages, we first standardize raw videos and crop patches that contain the hand regions.
As shown in Fig.~\ref{fig:data_preprocessing}, the input to the pipeline is a sign video with any dimensions.
First, a detection model~\cite{lyu2022rtmdetempiricalstudydesigning} predicts the signer's bounding box in all frames, and then the frames are cropped.
The output of this step is a cropped video with the same height and width.
Next, we leverage a pretrained body segmentation model~\cite{liu2024explorehumanparsingmodality} and a keypoint detection model~\cite{khirodkar2024sapiensfoundationhumanvision} to extract body segments and keypoints separately.
We noticed that some frames at the beginning and end of most videos do not contain useful information because both hands are hanging naturally or do not exist.
Therefore, we remove those frames to improve the efficiency of the mask-and-reconstruct pretraining.
Specifically, we check the following parameters to determine whether an arm hangs naturally in a given frame: 1) Angle $a_1$ formed by the left shoulder, right shoulder, and elbow; 2) Angle $a_2$ formed by the shoulder, elbow, and wrist on the same side of the body; 3) Distance $d_1$ between the shoulder and elbow on the same side of the body; and 4) Distance $d_2$ between the elbow and wrist on the same side of the body.
If $a_1$ is close to $90^\circ$, $a_2$ is close to $180^\circ$, and $d_1$ and $d_2$ are similar, the corresponding arm will be considered hanging naturally.
If both arms are hanging naturally, or both hands are not present in the frame, the frame is removed from the input.
Finally, the video is divided into small patches of size 2 $\times$ 16 $\times$ 16 and important patches that contain hands are extracted.

\begin{figure*}[t]
\centering
\includegraphics[width=1\textwidth]{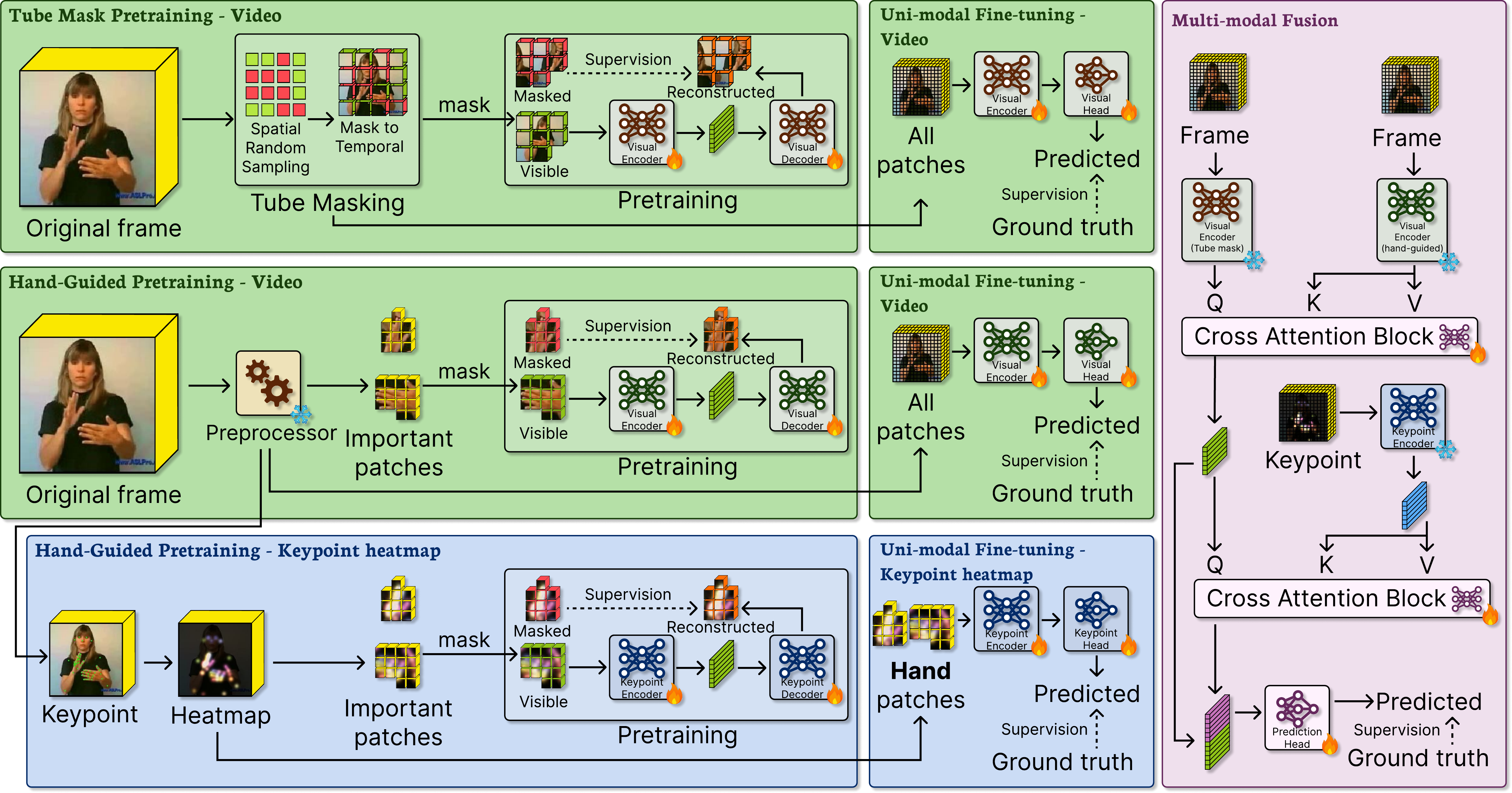}
\caption{Illustration of our proposed SignMAE framework. The framework contains hand-guided pretraining, uni-modal finetuning, and multi-modal fusion. For hand-guided pretraining, we designed spatio-temporal masking to learn local hand characteristics and leverage tube masking to learn global aspects, separately. For the downstream ISLR task, we first finetune two video encoders and one keypoint heatmap encoder separately. Next, we freeze the three uni-modal encoders and fuse the representation with cross-attention modules. The output representation from the video-keypoint cross‑attention module is concatenated with the representations from the video cross‑attention module. The concatenated feature vector is finally fed to a prediction head to perform recognition.}
\label{fig:main}
\end{figure*}

\subsection{Multi-Stream Pretraining}
As shown in Fig.~\ref{fig:main}, our multi-stream encoder adopts a three-stream architecture in which each encoder is pretrained independently to specialize in different modalities and levels of abstraction.
Specifically, we employ:

\begin{itemize}
\item A video encoder pretrained using random tube masking, aimed at capturing coarse-grained global context and long-range motion patterns.
\item A video encoder pretrained using hand–arm spatio-temporal masking, designed to focus on linguistically informative regions such as hands and arms.
\item A keypoint heatmap encoder pretrained with the spatio-temporal masking that focuses on the hand region only, which operates on pose-level representations to emphasize motion information.
\end{itemize}

Each stream is pretrained separately without shared weights or continuous training across streams.
This design ensures that each encoder learns a complementary inductive bias: global dynamics, local hand–arm interactions, and keypoint motion patterns, respectively.
The pretrained encoders are later integrated during a fusion stage to form a unified representation for ISLR.

\subsubsection{Tube Masking Pretraining for Video.}
We pretrain the first video encoder using the standard tube masking approach adopted in VideoMAEv2~\cite{wang2023videomaev2scalingvideo}.
This strategy randomly masks tubes across the video sequence, encouraging the model to capture global patterns and long-range temporal dependencies.
Although tube masking does not prioritize semantically important regions, it allows the encoder to develop a coarse understanding of overall sign dynamics, motion trajectories, and background context.

\subsubsection{Hand-Guided Pretraining for Video.}
At this stage, we pretrain the second video encoder by reconstructing hand and arm regions that are deliberately masked according to moving hands.
This hand-guided objective encourages the network to concentrate on the hand and arm areas where the linguistic information resides.
As shown in Fig.~\ref{fig:main}, pretraining on sign language data is conducted by masking out hands and arms in the input and reconstructing the masked region.
We employ vanilla Vision Transformer~\cite{dosovitskiy2021imageworth16x16words} as the backbone for both video and keypoint heatmaps.
In the original formulation of VideoMAEv2~\cite{wang2023videomaev2scalingvideo}, the masking strategy for the encoder is random tube masking, which aims to prevent information leakage caused by temporal correlations.
However, random tube masking is a suboptimal masking strategy for sign language for the following three reasons: 1) Over half of the patches in each video consist of only background; 2) The decoder cannot reconstruct the entire video if the encoder processes only background patches; and 3) As a result, the model pays insufficient attention to the arm and hand regions, which are important for accurate recognition.
To improve the overall efficiency of self-supervised learning on sign language video data, we propose to use a segmentation-based spatio-temporal masking strategy.
Specifically, given a video and its important patches covering the left and right hands and arms, the mask generator will select a masking option based on the presence and movement of the hands.

\noindent\emph{Directional spatial masking for two‑handed signs.}
When both hands are active, the hand and arm regions may partially overlap.
To mitigate potential information leakage, the mask generator first computes the overlap ratio between these regions.
If the ratio exceeds 25$\%$, all overlapping and non-overlapping hand patches are gathered for each frame, after which a masking direction (top, bottom, left, or right) is randomly chosen.
Approximately half of the patches along that direction are masked, and the remaining patches are retained for reconstruction.
When the overlap is below 25$\%$, indicating that the hands are largely separated, the mask generator instead reserves one arm–hand pair on the same side, while excluding any overlapping patches.

\noindent\emph{Directional spatial masking for one‑handed signs.}
For signs involving only one moving hand, the two-handed masking strategy is ineffective: masking the static hand and arm provides little semantic challenge, while masking the moving hand and arm removes critical information needed for reconstruction.
To address this, the mask generator preserves all patches from the upper portion of the moving arm, along with half of the patches from the moving hand.
As in the two-handed case, a masking direction (top, bottom, left, or right) is randomly selected, and approximately half of the hand patches along that direction are masked.

\noindent\emph{Temporal masking for all signs.}
Because adjacent frames in sign language videos exhibit high redundancy, we apply temporal masking to both two-handed and one-handed signs.
Specifically, the mask generator removes all patches from consecutive frames in the middle of the sequence, while the remaining patches are fed into the encoder.

\subsubsection{Hand-Guided Pretraining for Keypoint Heatmaps.}
The keypoint encoder is pretrained separately using a hand-guided spatio-temporal masking strategy, similar to that of the second video encoder, but with a key distinction: only the hand region is considered, and the arm region is excluded.
This choice reflects the structure of keypoint heatmaps, which emphasize joints and trajectories rather than pixel appearance.
To construct the input, we generate hand heatmaps using pose estimation outputs, and then apply spatio-temporal masking based on hand movement.
For each instance, the mask generator determines whether the sign is one- or two-handed and applies directional spatial masking to the hand patches accordingly.
Temporal masking is also applied by dropping consecutive frames in the middle of the sequence, encouraging the encoder to infer hand trajectories from sparse temporal cues.

\subsubsection{Mask Ratio Alignment.}
The encoder processes these reserved patches to produce a representation that serves as a prompt for the reconstruction process.
A key challenge in applying this segmentation-based masking strategy is maintaining the desired masking ratio for each sample.
To address this, we adjust the number of reserved patches to align with the masking ratio threshold.
If the masking ratio is lower than the threshold, the mask generator randomly masks additional patches at the edges of the reserved regions.
In contrast, if the masking ratio exceeds the threshold, the mask generator randomly unmasks patches adjacent to the reserved regions.
Once the masking ratio is aligned with the threshold, the patches covering the reserved body parts are retained and used by the encoder.
Next, the decoder takes the latent representations from the encoder and learnable mask tokens as inputs and reconstructs masked tokens.
Guided by the decoder mask, the video decoder only reconstructs hand and arm regions, and the keypoint decoder only reconstructs hand regions.
The model is trained with a mean squared error loss based only on masked patches:
\begin{equation}
\label{eq:mse_loss}
\mathcal{L} = \frac{1}{\Omega}\sum_{p \in \Omega} \lvert I(p) - \hat{I}(p) \rvert^{2},
\end{equation}
where $I(p)$ denotes the ground-truth video or keypoint heatmap, $\hat{I}(p)$ denotes the reconstructed one at position $p$, and $\Omega$ is the set of masked positions.
This mask-and-reconstruct pretraining helps the network infer the handshape of the masked hand from other visible parts, thereby learning the mutual context and coordination between hands and arms.
By applying directional spatial masking, the model is encouraged to reason about local dependencies within partially visible hand–arm regions.
Temporal masking further compels the model to infer motion trajectories and dynamic transitions across sparsely sampled frames, enhancing its ability to model temporal continuity and changes.
Together, these masking strategies promote the learning of spatially and temporally grounded representations.

\subsection{Finetuning and Fusion}
Since video appearance and keypoint motion provide complementary cues, we first finetune each encoder separately to stabilize their feature spaces, then freeze them and learn a light cross-attention module for efficient fusion.
After pretraining, we remove the reconstruction decoder and attach a lightweight linear classifier.
The encoder now takes unmasked sign clips as input and produces an embedding for every patch.
Mean pooling is applied to form a single clip-level feature, which the linear layer converts into class logits.
During finetuning, we apply batch-mode mixup~\cite{zhang2018mixupempiricalriskminimization} to both the video and keypoint streams.
The mixup generates synthetic data with soft labels, and the soft target cross-entropy loss function is used to train the model:
\begin{equation}
\mathcal{L}_{\text{soft-CE}}(\tilde{\mathbf{y}}, \mathbf{p})
  = -\sum_{c=1}^{K} \tilde{y}_{c}\,\log p_{c},
\end{equation}
where $\tilde{\mathbf{y}}$ is the soft label derived from the mixup and $\mathbf{p}$ is the predicted probability vector.

We finetune the two video encoders and the keypoint heatmap encoder independently, and observe that the video-based models consistently achieve stronger performance.
For the video encoder, all patches contribute to the loss, whereas for the keypoint encoder, only patches covering the hand regions are used in loss computation.
To leverage the complementary strengths of both modalities, we introduce a fusion stage that combines appearance cues from video with precise motion cues from keypoints.
During fusion, all encoder weights are frozen.
A cross-attention module first fuses features from the two video encoders: tokens from the encoder pretrained with tube masking are used as queries, while tokens from the encoder pretrained with spatio-temporal masking serve as keys and values.
The resulting fused video representation is then further integrated with keypoint features through a second cross-attention module, where the fused video tokens act as queries and the keypoint heatmap tokens as keys and values.

The output of this video–keypoint cross-attention module is concatenated with the intermediate video fusion features, yielding a joint representation that captures visual, motion, and cross-modal context, combining both global semantics and fine-grained local details.
This concatenated feature vector is passed to a classification head for final gloss prediction.

\section{Experiments}
\label{sec:experiments}
\subsection{Datasets}
We conduct experiments on three public sign language datasets, WLASL~\cite{li2020wordlevelsignlanguage}, NMFs-CSL~\cite{Hu_2021}, and Slovo~\cite{Kapitanov_2023}.

\begin{itemize}
\item WLASL is a large American Sign Language dataset capturing a vocabulary of 2,000 glosses (words) and 21,083 video samples. It is a challenging dataset collected from web videos performed by 119 signers. It consists of 14,289, 3,916, and 2,878 samples in the training, development, and test sets, respectively. The dataset is imbalanced, and some videos with different signs share the same label.
\item NMFs-CSL is a challenging Chinese Sign Language dataset involving many confusing words caused by fine-grained cues. The entire dataset is divided into 25,608 and 6,402 samples for training and testing, respectively.
\item Slovo is a large Russian Sign Language dataset that contains 1,000 glosses. The entire dataset is divided into 15,300 and 5,100 samples for training and testing, respectively.
\end{itemize}

\subsection{Evaluation Metrics}
Following~\cite{zuo2023natural,Jiang_2021_CVPR}, we report per-instance and per-class accuracy, denoting the average accuracy over each instance and each class separately.
We report the Top-1 and Top-5 accuracy for both per-instance and per-class for WLASL.
Since NMFs-CSL contains the same number of samples for each class, we only report per-instance accuracy.

\subsection{Implementation Details}
\subsubsection{Model Initialization and Data Preparation.}
In our experiments, we use Vision Transformer (large) as the backbone for both the video and keypoint heatmap models.
We initialize the parameters of the video model using the checkpoint from VideoMAE~\cite{tong2022videomaemaskedautoencodersdataefficient}.
The backbone of the checkpoint is Vision Transformer (large) pretrained on the Kinetics-400~\cite{kay2017kineticshumanactionvideo} dataset.
The keypoint heatmap model is initialized randomly.
Since no ground truth keypoint data is available in sign language datasets, we use Sapiens~\cite{khirodkar2024sapiensfoundationhumanvision} for its high performance to extract 55 keypoints, 13 body keypoints, and 42 hand keypoints.
Next, the keypoint coordinates are converted to a keypoint heatmap with 224 $\times$ 224 resolution.
To improve efficiency, we pretrain separate encoders on the three language groups: the American Sign Language (ASL) dataset (WLASL), the Chinese Sign Language (CSL) dataset (NMFs-CSL), and the Russian Sign Language (RSL) dataset (Slovo).
During finetuning, we initialize each task with the weights that match its language family: ASL pretrained weights for WLASL, CSL pretrained weights for NMFs-CSL, and RSL pretrained weights for Slovo.

\subsubsection{Hyper-parameters.}
The models are pretrained with a batch size of 64 for 600 epochs.
We use an AdamW~\cite{loshchilov2019decoupledweightdecayregularization} optimizer in our experiments.
For pretraining, the $\beta1$, $\beta2$, $\epsilon$, and weight decay are set to 0.9, 0.95, 1 $\times$ $10^{-8}$, and 0.1, respectively.
The learning rates for the video and keypoint model are $5\times 10^{-5}$ and $4 \times 10^{-6}$.
For the downstream ISLR finetuning, the learning rate of the models is $1\times 10^{-3}$.
For the multi-modal fusion, the learning rate is $1\times 10^{-4}$.

\section{Results}
\label{sec:results}
\begin{table}[t]
\small
\centering
\caption{Results on the WLASL2000 dataset. Comparison with the state-of-the-art in terms of per-instance and per-class accuracy. $^\dagger$denotes methods using both 32-frame and 64-frame clips as input. $^*$denotes methods using more modalities than ours, e.g., optical flow, depth map, and depth flow.}
\begin{tabular}{l|cc|cc}
\toprule
& \multicolumn{2}{c|}{\textbf{Per-instance}} & \multicolumn{2}{c}{\textbf{Per-class}} \\
\cline{2-5}
\textbf{Method} & \textbf{Top-1} & \textbf{Top-5} & \textbf{Top-1} & \textbf{Top-5} \\
\midrule
% I3D~\cite{li2020wordlevelsignlanguage} & 32.48 & 57.31 & - & - \\
% ST-GCN~\cite{yan2018spatialtemporalgraphconvolutional} & 34.40 & 66.57 & 32.53 & 65.45 \\
% Fusion-3~\cite{Hosain_2021_WACV} & 38.84 & 67.58 & - & - \\
BSL (multi-crop)~\cite{albanie2021bsl1kscalingcoarticulatedsign} & 46.82 & 79.36 & 44.72 & 78.47 \\
SignBERT~\cite{hu2021signbert} & 47.76 & 83.32 & 45.17 & 82.32 \\
HMA~\cite{Hu_2021} & 51.39 & 86.34 & 48.75 & 85.74 \\
VSNet~\cite{Li_2025_CVPR} & 53.54 & 83.18 & 51.18 & 82.03 \\
BEST~\cite{zhao2023best} & 54.59 & 88.08 & 52.12 & 87.28 \\
MASA~\cite{zhao2024masamotionawaremaskedautoencoder} & 55.77 & 88.85 & 53.13 & 88.14 \\
VKNet$^\dagger$~\cite{zuo2023natural} & 57.19 & 88.29 & 54.35 & 87.49 \\
SAM-SLR-v2$^*$~\cite{jiang2021signlanguagerecognitionskeletonaware} & 59.39 & 91.48 & 56.63 & 90.89 \\
\midrule
SignMAE (ours) & \textbf{60.95} & \textbf{91.94} & \textbf{58.29} & \textbf{91.37} \\
\bottomrule
\end{tabular}
% \vspace{-3mm}
\label{table:WLASL_accuracy}
\end{table}

\begin{table*}[t]
\small
\centering
\caption{Results on the NMFs-CSL and Slovo datasets in terms of per-instance accuracy. $^\dagger$ denotes methods using both 32-frame and 64-frame clips as input.}
\label{table:nmfs_slovo_accuracy}
\begin{minipage}[t]{0.48\textwidth}
\centering
\textbf{(a) NMFs-CSL}
\vspace{1mm}

\begin{tabular}{l|cc}
\toprule
\textbf{Method} & \textbf{Top-1} & \textbf{Top-5} \\
\midrule
Slowfast~\cite{feichtenhofer2019slowfastnetworksvideorecognition} & 66.3 & 86.6 \\
GLE-Net~\cite{Hu_2021} & 69.0 & 88.1 \\
VSNet~\cite{Li_2025_CVPR} & 75.3 & 95.4 \\
HMA~\cite{Hu_Zhou_Li_2021} & 75.6 & 95.3 \\
SignBERT~\cite{hu2021signbert} & 78.4 & 97.3 \\
BEST~\cite{zhao2023best} & 79.2 & 97.1 \\
VKNet$^\dagger$~\cite{zuo2023natural} & 79.7 & 97.9 \\
MASA~\cite{zhao2024masamotionawaremaskedautoencoder} & 80.1 & 98.3 \\
\midrule
SignMAE (ours) & \textbf{80.8} & \textbf{98.5} \\
\bottomrule
\end{tabular}
\end{minipage}
\hfill
\begin{minipage}[t]{0.48\textwidth}
\centering
\textbf{(b) Slovo}
\vspace{1mm}

\begin{tabular}{l|cc}
\toprule
\textbf{Method} & \textbf{Top-1} & \textbf{Top-5} \\
\midrule
Swin-large-48~\cite{liu2021videoswintransformer} & 55.66 & - \\
MViTv2-small-48~\cite{li2022mvitv2improvedmultiscalevision} & 62.18 & - \\
MViTv2-small-32~\cite{li2022mvitv2improvedmultiscalevision} & 64.09 & - \\
VSNet~\cite{Li_2025_CVPR} & 76.53 & 93.08 \\
\midrule
SignMAE (ours) & \textbf{76.86} & \textbf{95.23} \\
\bottomrule
\end{tabular}
\end{minipage}
\end{table*}

\subsection{Comparison with State-of-the-Art Encoders}
To ensure a fair comparison, we evaluated our approach against baselines that adopt a similar architectural paradigm.
Specifically, our method proposes a dedicated sign language encoder rather than a full ``encoder + classifier'' pipeline.
Accordingly, we only compare with baselines that consist of an encoder followed by a simple linear classification head.
Methods that rely on different configurations, such as task-specific heads or additional modules, are excluded, as their objectives and learning dynamics are not directly comparable to our encoder-focused design.

Table~\ref{table:WLASL_accuracy} summarizes the results for the WLASL2000 dataset.
The prior state-of-the-art method, SAM‑SLR-v2~\cite{jiang2021signlanguagerecognitionskeletonaware}, contains a heavyweight multi-modal encoder that jointly processes video frames, skeletal keypoints, optical flow, depth maps, and depth flow.
VKNet, the sign language encoder in NLA-SLR~\cite{zuo2023natural}, trims the input down to video frames and keypoint heatmaps, but still requires 64‑frame clips for each modality.
Our encoder likewise uses only video frames and keypoint heatmaps, yet operates on clips of just 32 frames, halving the temporal input.
Despite this lighter configuration, it improves on SAM‑SLR-v2 by 1.56$\%$ in per‑instance Top‑1 accuracy and by 1.66$\%$ in per‑class Top‑1 accuracy.

As shown in Table~\ref{table:nmfs_slovo_accuracy}(a), our encoder outperforms the previous best encoder MASA by 0.7$\%$ for Top-1 accuracy and by 0.2$\%$ for Top-5 accuracy on the NMFs-CSL dataset.

Finally, we evaluate our encoder on the Slovo dataset as shown in Table~\ref{table:nmfs_slovo_accuracy}(b).
Our encoder also outperforms the previous best encoder by 0.33$\%$ for Top-1 accuracy and by 2.15$\%$ for Top-5 accuracy.

\subsection{Ablation Study}
\label{sec:ablation_study}
\begin{table}[t]
\small
\centering
\caption{Ablation study on masking strategies. ST mask denotes a spatio-temporal hand-arm mask. Tube mask means a random tube mask. All results are based on the WLASL2000 dataset.}
\begin{tabular}{c|c|c|cc|cc}
\toprule
& \textbf{Encoder} & \textbf{Reconstruction} & \multicolumn{2}{c|}{\textbf{Per-instance}} & \multicolumn{2}{c}{\textbf{Per-class}} \\ \cline{4-7}
\textbf{Modality} & \textbf{Mask} & \textbf{Region} & \textbf{Top-1} & \textbf{Top-5} & \textbf{Top-1} & \textbf{Top-5} \\
\midrule
\multirow{4}{*}{Video} &
Random & Whole frame & 52.50 & 86.38 & 49.44 & 85.12 \\
& Tube & Whole frame & 55.45 & \textbf{87.63} & 52.51 & 85.37 \\
& ST mask & Whole frame & 52.19 & 85.30 & 49.39 & 83.82 \\
& ST mask & Hand and arm region & \textbf{55.77} & 86.90 & \textbf{53.24} & \textbf{85.49} \\
\midrule
& Random & Whole frame & 24.36 & 55.14 & 22.84 & 52.57 \\
Keypoint & Tube & Whole frame & 26.48 & 56.88 & 23.13 & 54.92 \\
Heatmap & ST mask & Whole frame & 40.31 & 71.40 & 38.26 & 69.95 \\
& ST mask & Hand region & \textbf{47.08} & \textbf{81.45} & \textbf{45.24} & \textbf{80.21} \\
\bottomrule
\end{tabular}
\label{table:masking_strategy}
\end{table}

\begin{table*}[t]
\small
\centering
\caption{Ablation studies on pretraining data scale and cross-attention layers}
\label{table:ablation_combined}

\begin{minipage}[t]{0.49\textwidth}
\centering
\textbf{(a) Pretraining data ratio (\%R)}\\[1mm]
\begin{tabular}{l|l|cc|cc}
\toprule
& & \multicolumn{2}{c|}{\textbf{Per-instance}} & \multicolumn{2}{c}{\textbf{Per-class}} \\
\cline{3-6}
\textbf{Modality} & \textbf{\%R} & \textbf{Top-1} & \textbf{Top-5} & \textbf{Top-1} & \textbf{Top-5} \\
\midrule
\multirow{5}{*}{Video}
& 0   & 48.51 & 82.73 & 44.34 & 81.17 \\
& 25  & 51.46 & 85.75 & 48.69 & 84.69 \\
& 50  & 52.02 & 85.34 & 49.15 & 83.92 \\
& 75  & 54.17 & 86.28 & 51.62 & 84.83 \\
& 100 & \textbf{55.77} & \textbf{86.90} & \textbf{53.24} & \textbf{85.49} \\
\midrule
\multirow{5}{*}{Keypoints}
& 0   & 3.27 & 11.43 & 2.17 & 8.75 \\
& 25  & 38.12 & 70.47 & 35.60 & 68.59 \\
& 50  & 41.38 & 72.48 & 37.14 & 70.93 \\
& 75  & 44.54 & 76.89 & 41.23 & 74.31 \\
& 100 & \textbf{47.08} & \textbf{81.45} & \textbf{45.24} & \textbf{80.21} \\
\bottomrule
\end{tabular}
\end{minipage}
\hfill
\begin{minipage}[t]{0.49\textwidth}
\centering
\textbf{(b) Cross-attention layers (\#L)}\\[1mm]
\begin{tabular}{l|cc|cc}
\toprule
& \multicolumn{2}{c|}{\textbf{Per-instance}} & \multicolumn{2}{c}{\textbf{Per-class}} \\
\cline{2-5}
\textbf{\#L} & \textbf{Top-1} & \textbf{Top-5} & \textbf{Top-1} & \textbf{Top-5} \\
\midrule
1 & 57.68 & 89.61 & 53.64 & 87.50 \\
2 & 59.38 & 90.90 & 55.05 & 89.49 \\
4 & \textbf{60.95} & \textbf{91.94} & \textbf{58.29} & \textbf{91.37} \\
8 & 59.66 & 91.00 & 55.71 & 89.54 \\
\bottomrule
\end{tabular}
\end{minipage}
\end{table*}

\begin{table}[t]
\small
\centering
\caption{Ablation study on the multi-modal fusion.}
\begin{tabular}{l|l|cc|cc}
\toprule
\multirow{3}{*}{\textbf{Dataset}} & \multirow{3}{*}{\textbf{Modality}} & \multicolumn{2}{c|}{\textbf{Per-instance}} & \multicolumn{2}{c}{\textbf{Per-class}} \\
\cline{3-6}
& & \textbf{Top-1} & \textbf{Top-5} & \textbf{Top-1} & \textbf{Top-5} \\
\midrule
\multirow{5}{*}{WLASL2000}
& Video (ST mask) & 55.77 & 86.90 & 53.24 & 85.49 \\
& Video (Tube mask) & 55.45 & 87.63 & 52.51 & 85.37 \\
& Keypoint & 47.08 & 81.45 & 45.24 & 80.21 \\
& Video-Video Fusion & 58.72 & 90.93 & 56.03 & 90.27 \\
& Video-Keypoint Fusion & \textbf{60.95} & \textbf{91.94} & \textbf{58.29} & \textbf{91.37} \\
\midrule
\multirow{3}{*}{NMFs-CSL}
& Video (ST mask) & 77.3 & 95.6 & - & - \\
& Video (Tube mask) & 76.8 & 95.5 & - & - \\
& Keypoint & 66.6 & 95.3 & - & - \\
& Video-Video Fusion & 77.5 & 95.6 & - & - \\
& Video-Keypoint Fusion & \textbf{80.8} & \textbf{98.5} & - & - \\
\bottomrule
\end{tabular}
\label{table:fusion}
\end{table}

\subsubsection{Masking Strategy.}
As shown in Table~\ref{table:masking_strategy}, we evaluate four masking strategies on the video stream and keypoint heatmap stream using the WLASL dataset.
In previous research, random encoder mask, random tube encoder mask, and running cell decoder mask were proposed for pretraining on generic action recognition datasets.
The first two schemes replicate VideoMAE's designs: a random patch mask that independently hides 90$\%$ of patches, and a random tube mask that removes or retains entire tubes (patches across all frames), requiring full-frame reconstruction in both cases.
VideoMAEv2 further lightens the decoder by passing it only a regularly spaced 50$\%$ subset of visible tokens (``running-cell'' masking), yet still expects a whole-frame reconstruction.
Our approach adopts a spatio-temporal hand-arm mask, revealing only patches that intersect detected hand patches, and constrains the decoder to reconstruct the corresponding hand region.
This focus-aligned strategy delivers the best results, lifting per-instance Top-1 accuracy on video from 55.45$\%$ under the random-tube baseline to 55.77$\%$ and on keypoints from 26.48$\%$ to 47.08$\%$.
When the spatio-temporal hand-arm mask is paired with whole-frame reconstruction, performance drops markedly (e.g., to 52.19$\%$ Top-1 on video), confirming that the reconstruction target must match the masked context for maximal benefit.

\subsubsection{Pretraining Data Scale.}
As shown in Table~\ref{table:ablation_combined}(a), as the ratio of pretraining data volume increases, the performance on the downstream ISLR task gradually increases on accuracy metrics.
This indicates that our proposed framework may benefit from larger pretraining datasets.

\subsubsection{Cross-Attention Layers.}
We conduct multi-modal fusion experiments with different numbers of cross-attention layers.
As shown in Table~\ref{table:ablation_combined}(b), the accuracy increases when the number of cross-attention layers increases.
The performance reaches the peak when there are 4 cross-attention layers.
Unless stated, we utilize 4 cross-attention layers in all our multi-modal fusion experiments.

\subsubsection{Multi-Modal Fusion.}
As shown in Table~\ref{table:fusion}, we evaluate the effectiveness of the multi-modal fusion with cross-attention on both WLASL and NMFs-CSL.
Across all benchmarks, the fusion of the two video streams consistently outperforms any single-stream video encoder.
Moreover, integrating video with keypoint heatmaps achieves the highest recognition accuracy, underscoring that the three streams contribute complementary information that no single modality can fully capture in isolation.
On WLASL, fusion raises per-instance Top-1 accuracy to 60.95$\%$—a gain of 2.23$\%$ and 13.87$\%$ over the two-stream video encoder and keypoint-only encoder, respectively—and similarly improves per-class Top-1 accuracy to 58.29$\%$.
On the more constrained NMFs-CSL dataset, fusion still enhances video baselines, reaching 80.8$\%$ on Top-1 accuracy and 98.5$\%$ on Top-5 accuracy, thereby outperforming the two-stream video encoder by 3.3$\%$ and widening the gap over keypoints to 14.2$\%$.
\vspace{-2mm}

\subsubsection{spatio-temporal Hand-Arm Mask and Tube Mask.}
\begin{figure}[t]
\centering
\includegraphics[width=1\linewidth]{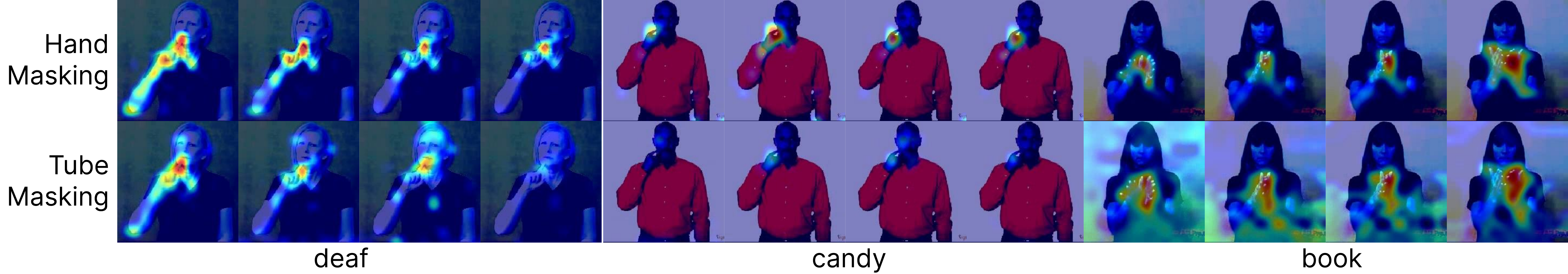}
\caption{Visualization of the attention from the encoder pre-trained with tube masking and spatio-temporal hand-arm masking.}
\label{fig:att_vis}
\end{figure}

Fig.~\ref{fig:att_vis} presents attention visualizations from encoders pretrained with tube masking and with spatio-temporal hand–arm masking.
The gloss \emph{book} is a two-handed sign involving coordinated motion of both hands.
While both models attend to the hand regions, the encoder pretrained with tube masking also allocates substantial attention to the body and background, whereas the spatio-temporal hand–arm mask encoder focuses more selectively on the hands.
For the one-handed sign \emph{candy}, where the hand remains relatively static, the tube-mask encoder fails to consistently capture the hand region, in contrast to the spatio-temporal hand–arm mask encoder, which reliably attends to the hand.
Similarly, for the one-handed sign \emph{deaf}, in which the hand moves around the face, the tube masking encoder disperses attention to surrounding regions such as the hair and face, while the spatio-temporal hand–arm mask encoder maintains concentrated attention on the moving hand.
Overall, the visualizations suggest that spatio-temporal hand–arm masking guides the model to attend more faithfully to the hands across both static and dynamic signing contexts.
In general, tube masking encourages attention to global aspects of the scene, whereas spatio-temporal hand–arm masking emphasizes local hand characteristics.
\vspace{-2mm}

\section{Conclusion}
\label{sec:conclusion}

In this paper, we have presented a four-stage pipeline that combines hand-guided self-supervised learning with cross-modal fusion to tackle isolated sign language recognition.
By explicitly locating hand regions during data preprocessing and guiding the mask-and-reconstruct pretraining with segmentation cues, our method overcomes the inefficiencies of random tube masking and learns representations that capture the fine-grained appearance and motion patterns unique to sign language.
Separate fine-tuning of the video and keypoint streams confirms the strength of the video encoder, while the ensuing cross-attention fusion demonstrates that complementary motion information can be harnessed without retraining either backbone.
Extensive experiments on large-scale benchmarks show consistent improvements over uni-modal baselines and prior state-of-the-art methods, validating both the proposed masking strategy and the lightweight fusion design. 

\subsubsection{Acknowledgment}
This research was partly supported by the Australian Government through the Australian Research Council's Discovery Early Career Researcher Award (project DE230100049).
The views expressed herein are those of the authors and are not necessarily those of the Australian Government or Australian Research Council.
We also acknowledge Monash University and National Computational Infrastructure for providing the High Performance Computing infrastructure used in this research.

\bibliographystyle{splncs04}
\bibliography{main.bib}
\end{document}